\documentclass{article}
\usepackage{spconf,amsmath,graphicx}
\usepackage{tipa}
\usepackage{CJKutf8}

\title{A Hybrid Text Normalization System using Multi-Head Self-Attention for Mandarin}
\name{Junhui Zhang, Junjie Pan, Xiang Yin, Chen Li, Shichao Liu, Yang Zhang, Yuxuan Wang, Zejun Ma}
\address{ByteDance AI-Lab\\\\\tt\small{\{zhangjunhui.915,panjunjie.jeff,yinxiang.stephen,
lichen.cherlyn,}\\\tt\small{liushichao,zhangyang.elfin,wangyuxuan.11,mazejun\}@bytedance.com}}
\begin{document}
%
\maketitle
\begin{abstract}
In this paper, we propose a hybrid text normalization system using multi-head self-attention. The system combines the advantages of a rule-based model and a neural model for text preprocessing tasks. Previous studies in Mandarin text normalization usually use a set of hand-written rules, which are hard to improve on general cases. The idea of our proposed system is motivated by the neural models from recent studies and has a better performance on our internal news corpus. This paper also includes different attempts to deal with imbalanced pattern distribution of the dataset. Overall, the performance of the system is improved by over 1.9\% on sentence-level. This idea can potentially be adopted by different languages with rule-based text normalization systems.

\end{abstract}
\section{Introduction}
\label{sec:intro}

Text Normalization (TN) is a process to transform non-standard words (NSW) into spoken-form words (SFW) for disambiguation. In Text-To-Speech (TTS), text normalization is an essential procedure to normalize unreadable numbers, symbols or characters, such as transforming ``\$20'' to ``twenty dollars'' and ``@'' to ``at'', into words that can be used in speech synthesis. The surrounding context is the determinant for ambiguous cases in TN. For example, the context will decide whether to read ``2019'' as year or a number, and whether to read ``10:30'' as time or the score of a game. In Mandarin, some cases depend on language habit instead of rules- ``2'' can either be read as ``\textipa{\`{e}}r'' or ``li\textipa{\v{a}}ng'' and ``1'' as ``y\textipa{\={\i}}'' or ``y\textipa{\={a}}o''.

Currently, based on the traditional taxonomy approach for NSW\cite{R11}, the Mandarin TN tasks are generally resolved by rule-based systems which use keywords and regular expressions to determine the SFW of ambiguous words\cite{R1,R3}. These systems typically classify NSW into different pattern groups, such as abbreviations, numbers, etc., and then into sub-groups, such as phone number, year, etc., which has corresponding NSW-SFW transformations. Zhou\cite{R5} and Jia\cite{R4} proposed systems which use maximum entropy (ME) to further disambiguate the NSW with multiple pattern matches. For the NSW given the context constraints, the highest probability corresponds to the highest entropy. Liou\cite{R2} proposed a system of data-driven models which combines a rule-based and a keyword-based TN module. The second module classifies preceding and following words around the keywords and then trains a CRF model to predict the NSW patterns based on the classification results. There are some other hybrid systems\cite{R13,R12} which use NLP models and rules separately to help normalize hard cases in TN.

For recent NLP studies, sequence-to-sequence (seq2seq) models have achieved impressive progress in TN tasks in English and Russian\cite{R6,R7}. Seq2seq models typically encode sequences into a state vector, which is decoded into an output vector from its learnt vector representation and then to a sequence. Different seq2seq models with bi-LSTM, bi-GRU with attention are proposed in \cite{R7,R8}. Zhang and Sproat proposed a contextual seq2seq model, which uses a sliding-window and RNN with attention\cite{R6}. In this model, bi-directional GRU is used in both encoder and decoder, and the context words are labeled with ``$\langle$self$\rangle$'', helping the model distinguish the NSW and the context.

However, seq2seq models have several downsides when directly applied in Mandarin TN tasks. As mentioned in \cite{R6}, the sequence output directly from a seq2seq model can lead to unrecoverable errors. The model sometimes changes the context words which should be kept the same. Our experiments produce similar errors. For example, ``Podnieks, Andrew 2000'' is transformed to ``Podncourt, Andrew Two Thousand'', changing ``Podnieks'' to ``Podncourt''. These errors cannot be detected by the model itself. In \cite{R14}, rules are applied to two specific categories to resolve silly errors, but this method is hard to apply to all cases. Another challenge in Mandarin is the word segmentation since words are not separated by spaces and the segmentation could depend on the context. Besides, some NSW may have more than one SFW in Mandarin, making the seq2seq model hard to train. For example, ``\begin{CJK*}{UTF8}{gbsn}两千零八年\end{CJK*}'' and ``\begin{CJK*}{UTF8}{gbsn}二零零八年\end{CJK*}'' are both acceptable SFW for ``2008\begin{CJK*}{UTF8}{gbsn}年\end{CJK*}''. The motivation of this paper is to combine the advantages of a rule-based model for its flexibility and a neural model to enhance the performance on more general cases. To avoid the problems of seq2seq models, we consider the TN task as a multi-class classification problem with carefully designed patterns for the neural model.

The contributions of this paper include the following. First, this is the first known TN system for Mandarin which uses a neural model with multi-head self-attention. Second, we propose a hybrid system combining a rule-based model and a neural model. Third, we experiment with different approaches to deal with imbalanced dataset in the TN task.

The paper is organized as follows. Section \ref{sec:method} introduces the detailed structure of the proposed hybrid system and its training and inference. In Section \ref{sec:experiments}, the performance of different system configurations is evaluated on different datasets. And the conclusion is given in Section \ref{sec:conclusion}.

\section{Method}
\label{sec:method}

\subsection{Rule-based TN model}
\label{subsec:rule}
The rule-based TN model can handle the TN task alone and is the baseline in our experiments. It has the same idea as in \cite{R6} but has a more complicated system of rules with priorities. The model contains 45 different groups and about 300 patterns as sub-groups, each of which uses a keyword with regular expressions to match the preceding and following texts. Each pattern also has a priority value. During normalization, each sentence is fed as input and the NSW will be matched by the regular expressions. The model tries to match patterns with longer context and slowly decrease the context length until a match is found. If there are multiple pattern matches with the same length, the one with a higher priority will be chosen for the NSW. The model has been developed on abundant test data and bad cases. The advantage of the rule-based system is the flexibility, since one can simply add more special cases when they appear, such as new units. However, improving the performance of this system on more general cases becomes a bottleneck. For example, in a report of a football game, it cannot transform ``1-3'' to score if there are no keywords like ``score'' or ``game'' close to it.

\subsection{Proposed Hybrid TN system}
\label{subsec:system}
\begin{figure}[t]
\centering
\includegraphics[width=7.5cm]{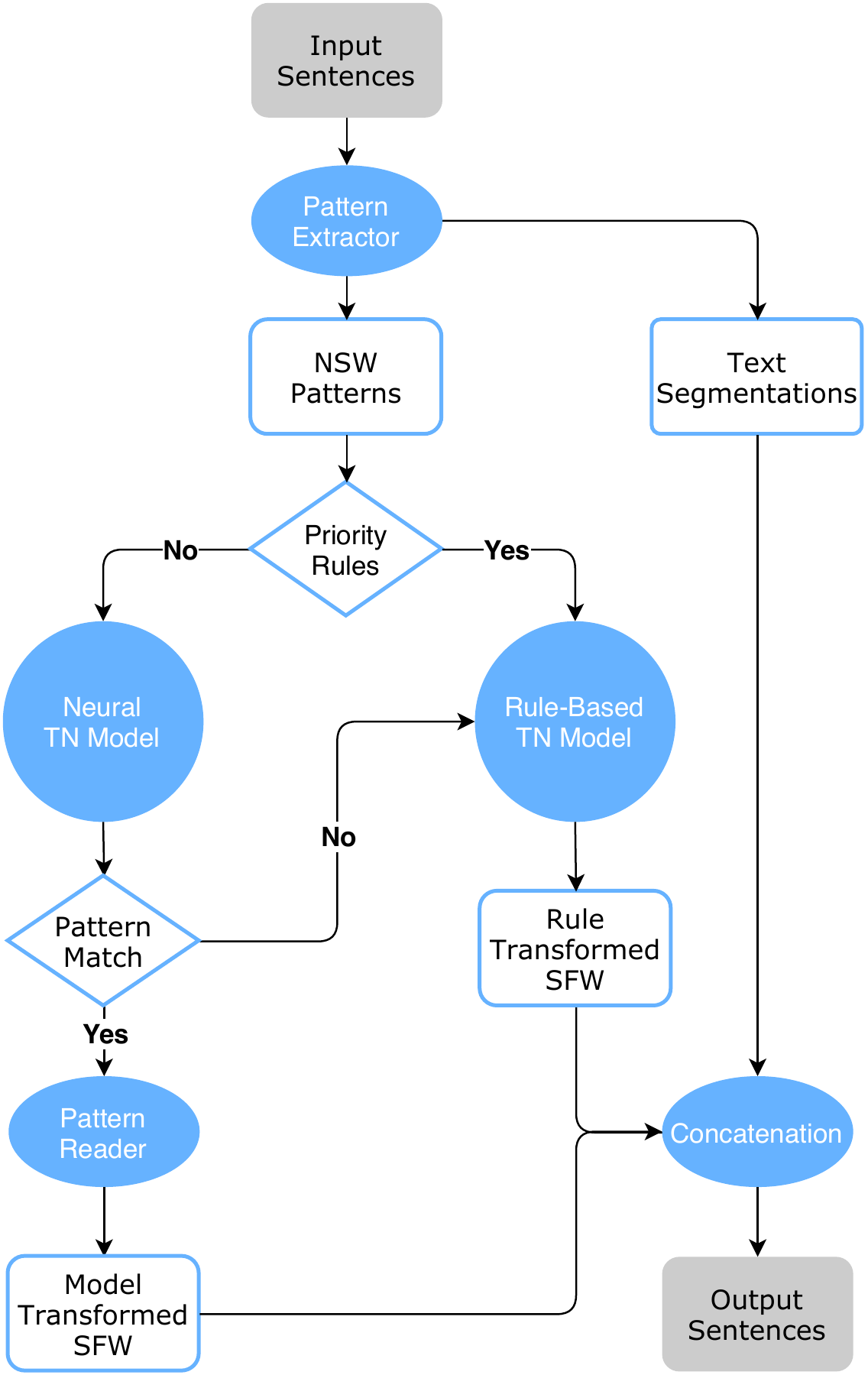}
\caption{Flowchart of the proposed hybrid TN system.}
\label{fig:pipeline}
\end{figure}

We propose a hybrid TN system as in Fig. \ref{fig:pipeline}, which combines the rule-based model and a neural model. The NSW are first extracted from the input text using regular expressions. We only extract NSW that are digit and symbol related, and other NSW like abbreviations will be processed in the rule-based model. Then the system performs a priority check on the NSW, and the matched NSW will be sent into the rule-based model. The priority rules include definite NSW such as ``911'' and user-defined strings. All of the remaining patterns are passed through the neural model to be classified into one of the pattern groups. Before normalizing the classified NSW in the pattern reader, the format of each classified NSW is checked with regular expressions, and the illegal ones, such as classifying ``10\%'' to read as year, will be filtered back to the rule-based system. In the pattern reader, each pattern label has a unique process function to perform the NSW-SFW transformation. Finally, all of the normalized SFW are inserted back to the text segmentations to form the output sentences. For the entire system, the neural model serves the major role. In our golden set test, the priority rules filter 22.8\% of all patterns while the neural model handles 77.8\%, 2.2\% of which fail the pattern match and flow back to the rule-based model.

\begin{figure}[t]
\centering
\includegraphics[width=7cm]{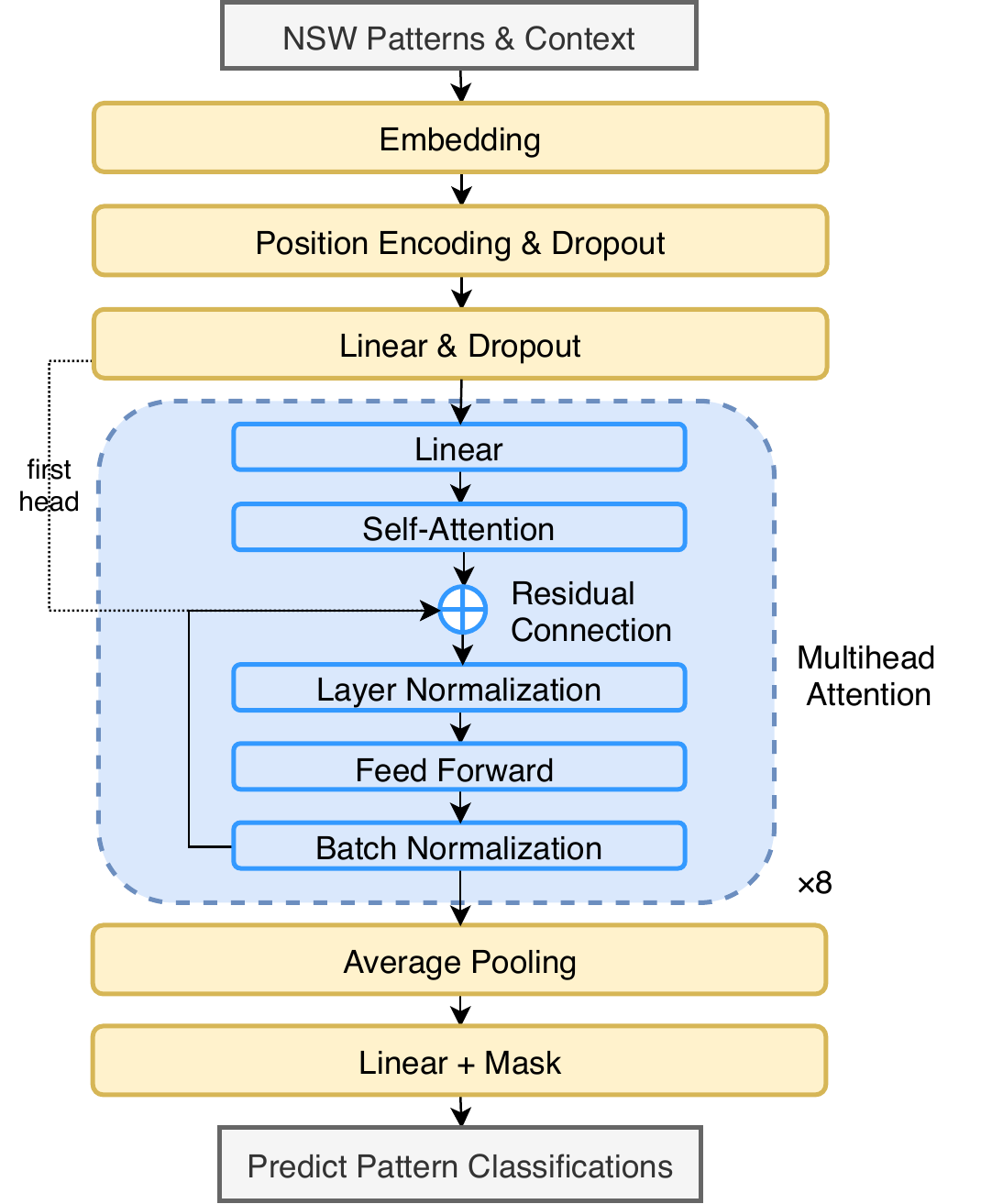}
\caption{Multi-head self-attention model structure.}
\label{fig:model}
\end{figure}

Multi-head self-attention was proposed in transformer\cite{R9}, which uses self-attention in the encoder and decoder and encoder-decoder attention in between. Motivated by this structure, multi-head self-attention is adopted in our neural model and the structure is shown in Fig. \ref{fig:model}. Compared with other modules like LSTM and GRU, self-attention can efficiently extract the information of the NSW with all context in parallel and is fast to train. The core part of the neural model is similar to the encoder of a transformer. The inputs of the model are the sentences with their manually labeled NSW. We take a 30-character context window around each NSW and send it to the embedding layer. Padding is used when the window exceeds the sentence range. After 8 heads of self-attention, the highest masked softmax probability is chosen as the classified pattern group. The mask uses a regular expression to check if the NSW contain symbols and filters illegal ones such as classifying ``12:00'' as pure number, which is like a bi-class classification before softmax is applied.

For the loss function, in order to solve the problem of imbalanced dataset, which will be talked about in \ref{subsec:dataset}, the final selection of the loss function is motivated by \cite{R10}:

\begin{equation} L= \left\{\begin{aligned}-\alpha_t(1-p)^{\gamma} \log (p), & \text { if } \quad y=1 \\-\alpha_t p^{\gamma} \log (1-p), & \text { if } \quad y=0 \end{aligned}\right.\end{equation}
where $\alpha_t$ and $\gamma$ are hyper-parameters, $p$'s are the pattern probabilities after softmax, and $y$ is the correctness of the prediction. In our experiment, we choose $\alpha_t=0.5$ and $\gamma=4$.

\subsection{Training and Inference}
\label{subsec:train}
The neural TN model is trained alone with inputs of labeled sentences and outputs of pattern groups. And the inference is on the entire hybrid TN system in Fig1, which takes the original text with NSW as input and text with SFW as output.

The training data is split into 36 different classes, each of which has its own NSW-SFW transformation. The distribution of the dataset is the same with the NSW in our internal news corpus and is imbalanced, which is one of the challenges for our neural model. The approaches to deal with the imbalanced dataset are discussed in the next section.

\section{Experiments}
\label{sec:experiments}

\subsection{Training Dataset}
\label{subsec:dataset}
\begin{figure}[htb]
\centering
\includegraphics[width=7cm]{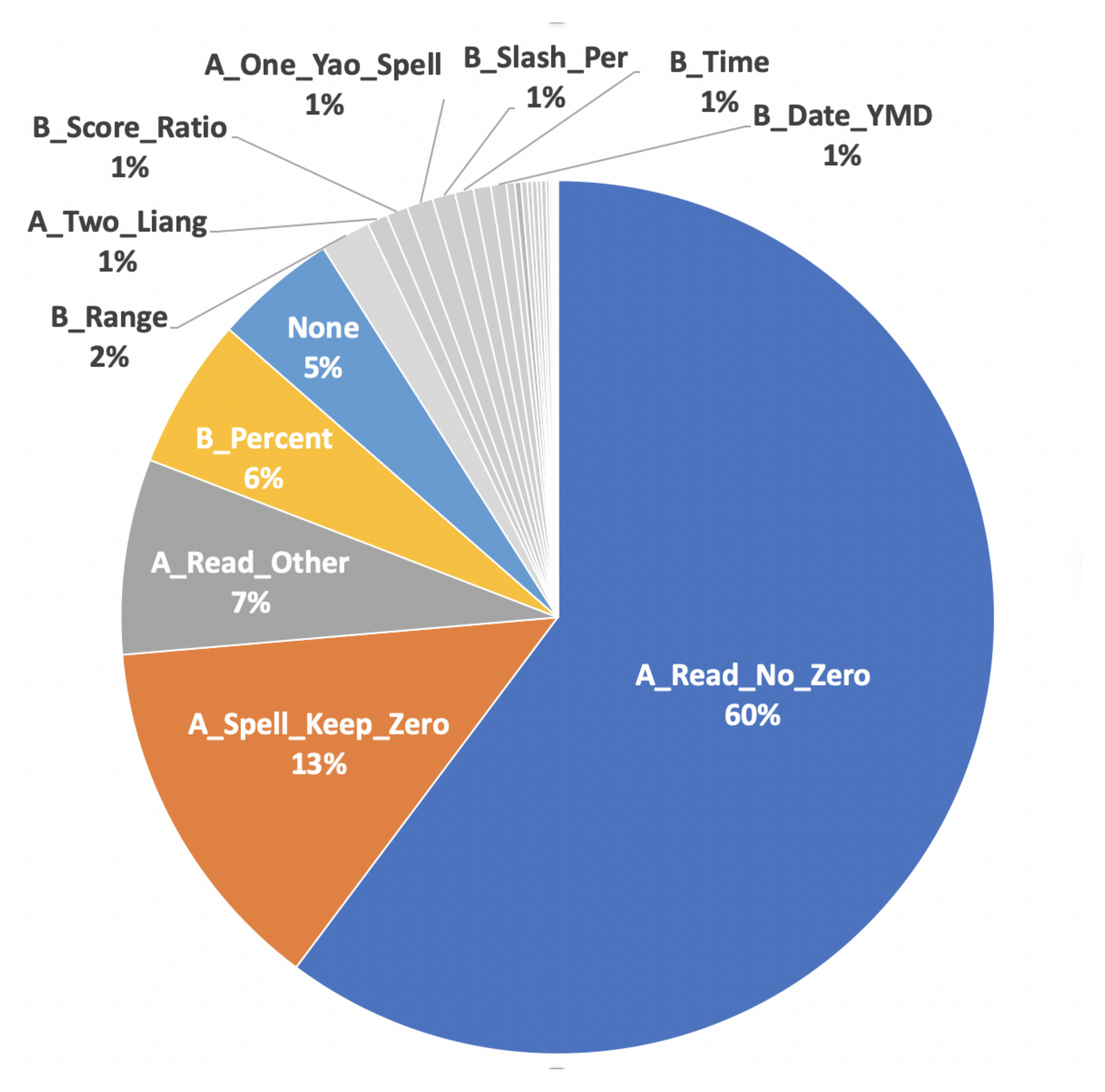}
\caption{Label distribution for dataset.}
\label{fig:label}
\end{figure}

The training dataset contains 100,747 pattern labels. The texts are in Mandarin with a few English words. The patterns are digit or symbol related, and patterns like abbreviations are not included. There are 36 classes in total, and some examples are listed in Table \ref{ta:dataset}. The first 8 are patterns with digits and symbols, and there could be substitutions among ``$\sim$'', ``-'', ``—'' and ``:'' in a single group. The last 2 are language related- ``1'' and ``2'' have different pronunciations based on language habit in Mandarin. Fig. \ref{fig:label} is a pie chart of the training label distribution. Notice that the top 5 patterns take up more than 90\% of all labels, which makes the dataset imbalanced.

\begin{table}[!ht]
\centering\small
\caption{Examples of some dataset pattern rules.}\label{ta:dataset}
\begin{tabular}{l|l}
\hline
Pattern Name&Pattern Example\\
\hline
A\_Read\_No\_Zero&\underline{200} people\\
A\_Spell\_Keep\_Zero&The \underline{2020} Conference\\
B\_Percent&Only \underline{10\%} of students voted\\
B\_Range&about \underline{10-15} degree\\
B\_Score\_Ratio&Team A is \underline{30-10} leading\\
B\_Slash\_Per&There are five people\underline{/}group\\
B\_Time&It starts at \underline{10:30}\\
B\_Date\_YMD&Today is \underline{2019-10-01}\\
\hline
A\_Two\_Liang&\underline{2}\begin{CJK*}{UTF8}{gbsn}个人\end{CJK*} (2 people)\\
A\_One\_Yao\_Spell&\begin{CJK*}{UTF8}{gbsn}打\end{CJK*}\underline{911} (Call 911)\\
\hline
\end{tabular}
\end{table}

Imbalanced dataset is a challenge for the task because the top patterns are taking too much attention so that most weights might be determined by the easier ones. We have tried different methods to deal with this problem. The first method is data expansion using oversampling. Attempts include duplicating the text with low pattern proportion, replacing first few characters with paddings in the window, randomly changing digits, and shifting the context window. The other method is to add loss control in the model as mentioned in \ref{subsec:system}. The loss function helps the model to focus on harder cases in different classes and therefore reduce the impact of the imbalanced data. The experimental results are in \ref{subsec:result}.

\subsection{System Configuration}
\label{subsec:config}
For sentence embedding, pre-trained embedding models are used to boost training. We experiment on a word-to-vector (w2v) model trained on Wikipedia corpus and fine-tuning a trained BERT\cite{R15} model. The experimental result is in \ref{subsec:result}.

The experiments show that using a fixed context window achieves better performance than padding to the maximum length of all sentences. And padding with 1's gives a slightly better performance than with 0's. During inference, all NSW patterns in one sentence need to be processed simultaneously before transforming to SFW to keep their original context.

\subsection{Model Performance}
\label{subsec:result}
Table \ref{ta:compare} compares the pattern accuracies on the test set with 7 different neural model setups. Model 2-7's configuration differences are compared with Model 1: \raisebox{.5pt}{\textcircled{\raisebox{-.9pt} {1}}} proposed configuration; \raisebox{.5pt}{\textcircled{\raisebox{-.9pt} {2}}} fine-tune with BERT; \raisebox{.5pt}{\textcircled{\raisebox{-.9pt} {3}}} replace padding with 1's with 0's; \raisebox{.5pt}{\textcircled{\raisebox{-1.1pt} {4}}} replace the context window length of 30 with maximum sentence length; \raisebox{.5pt}{\textcircled{\raisebox{-1.1pt} {5}}} replace the loss with Cross Entropy (CE) loss; \raisebox{.5pt}{\textcircled{\raisebox{-1.1pt} {6}}} remove mask; \raisebox{.5pt}{\textcircled{\raisebox{-1.0pt} {7}}} apply data expansion.

\begin{table}[!ht]
\centering\small
\caption{Comparison of different experimental setups.}\label{ta:compare}
\begin{tabular}{l|c}
\hline
Experimental setup&Accuracy\\
\hline
Model 1 (proposed)&0.916\\
Model 2 (+ BERT)&0.904\\
Model 3 (+ pad 0's)&0.914\\
Model 4 (+ max window)&0.907\\
Model 5 (+ CE loss)&0.913\\
Model 6 (- mask)&0.910\\
Model 7 (+ data expansion)&0.908\\
\hline
\end{tabular}
\end{table}

Overall, w2v model has a better performance than fine-tuning with BERT. Various BERT models are used but none of them beat the highest accuracy. A possible reason is that the model easily overfits the training data. It also shows that data expansion does not give better accuracy even though we find the model becomes more robust and has better performance on the lower proportioned patterns. This is because the pattern distribution changes and its performance on the top proportioned patterns decreases a little, resulting in a large number of misclassifications. This is a tradeoff between a robust and a high-accuracy model. We choose Model 1 for the following test since the golden set is evaluated by accuracy.

The neural model with the proposed configuration is evaluated on the test set of each pattern group using precision, recall and $F_1$ score, which is the harmonic mean of precision and recall. The results of the top proportioned patterns are shown in Table \ref{ta:test}. This result can help determine which well-predicted patterns to be used from the neural model.



\begin{table}[!ht]
\centering\small
\caption{Model performance on the test dataset.}\label{ta:test}
\begin{tabular}{l|c|c|c}
\hline
Pattern Name&Precision&Recall&$F_1$\\
\hline
A\_Read\_No\_Zero&0.974&0.979&0.977\\
A\_Spell\_Keep\_Zero&0.932&0.916&0.924\\
B\_Percent&0.998&0.990&0.994\\
B\_Range&0.932&0.932&0.932\\
B\_Time&0.969&0.912&0.939\\
B\_Score\_Ratio&0.962&0.962&0.962\\
B\_Slash\_Per&0.994&0.966&0.980\\
B\_Date\_YMD&1.000&0.923&0.960\\
A\_Two\_Liang&0.613&0.797&0.693\\
A\_One\_Yao\_Spell&0.637&0.631&0.634\\
\hline
\multicolumn{2}{c|}{Overall Accuracy}&\multicolumn{2}{c}{0.916}\\
\hline
\end{tabular}
\end{table}

The proposed hybrid TN system is tested on an internal golden set of NSW-SFW pairs. It would be considered as an error if any character in the transformed and ground-truth sentences is different. The golden set has 67853 sentences, each of which contains 1-10 NSW strings. The sentence and average pattern accuracies are listed in Table \ref{ta:goldenset}. On sentence-level, the accuracy increases by 1.9\%, which indicates an improvement of correctness on over 1000 sentences. The improvement is mainly on ambiguous NSW with few keywords around. The average accuracy of the hybrid system is also higher than the pure data-driven neural model from Table \ref{ta:compare}.

\begin{table}[!ht]
\centering\small
\caption{Model performance on the news golden set.}\label{ta:goldenset}
\begin{tabular}{c|c|c}
\hline
&Sentence Accuracy&Pattern Accuracy\\
\hline
Rule-based TN model&0.867&0.946\\
Proposed TN system&0.886&0.955\\
\hline
\end{tabular}
\end{table}

\section{Conclusions \& Future Work}
\label{sec:conclusion}

In this paper, we propose a hybrid TN system for Mandarin using multi-head self-attention. This system aims to dealing with the bottleneck of the performance of a highly developed rule-based model with the advantages of a neural model. The system mainly relies on the neural model instead of rules. From the test results, the proposed system improves the accuracy on NSW-SFW transformation by over 1.9\% on sentence-level and still has a potential to improve further. The hybrid system structure can be beneficial to other languages with TN rules, and help increase their generalization.

The future work includes other aspects of model explorations. Tokenization for Mandarin will be applied to replace the character-wise embedding with word-level embedding. Seq2seq models will be applied to help replace the rules with an end-to-end system. And more labeled dataset in other corpus will be supplemented for training and evaluation.

\bibliographystyle{IEEEbib}
\bibliography{refs}

\begin{thebibliography}{10}

\bibitem{R11}
Richard Sproat, Alan~W Black, Stanley Chen, Shankar Kumar, Mari Ostendorf, and
  Christopher Richards,
\newblock ``Normalization of non-standard words,''
\newblock {\em Computer speech \& language}, vol. 15, no. 3, pp. 287--333,
  2001.

\bibitem{R1}
Sunhee Kim,
\newblock ``Corpus-based evaluation of chinese text normalization,''
\newblock in {\em 2017 20th Conference of the Oriental Chapter of the
  International Coordinating Committee on Speech Databases and Speech I/O
  Systems and Assessment (O-COCOSDA)}. IEEE, 2017, pp. 1--4.

\bibitem{R3}
Xinxin Zhou, Zhiyong Wu, Chun Yuan, and Yuzhuo Zhong,
\newblock ``Document structure analysis and text normalization for chinese
  putonghua and cantonese text-to-speech synthesis,''
\newblock in {\em 2008 Second International Symposium on Intelligent
  Information Technology Application}. IEEE, 2008, vol.~1, pp. 477--481.

\bibitem{R5}
Yuxiang Jia, Dezhi Huang, Wu~Liu, Yuan Dong, Shiwen Yu, and Haila Wang,
\newblock ``Text normalization in mandarin text-to-speech system,''
\newblock in {\em 2008 IEEE International Conference on Acoustics, Speech and
  Signal Processing}. IEEE, 2008, pp. 4693--4696.

\bibitem{R4}
Tao Zhou, Yuan Dong, Dezhi Huang, Wu~Liu, and Haila Wang,
\newblock ``A three-stage text normalization strategy for mandarin
  text-to-speech systems,''
\newblock in {\em 2008 6th International Symposium on Chinese Spoken Language
  Processing}. IEEE, 2008, pp. 1--4.

\bibitem{R2}
Guan-Ting Liou, Yih-Ru Wang, and Chen-Yu Chiang,
\newblock ``Text normalization for mandarin tts by using keyword information,''
\newblock in {\em 2016 Conference of The Oriental Chapter of International
  Committee for Coordination and Standardization of Speech Databases and
  Assessment Techniques (O-COCOSDA)}. IEEE, 2016, pp. 73--78.

\bibitem{R13}
Richard Beaufort, Sophie Roekhaut, Louise-Am{\'e}lie Cougnon, and C{\'e}drick
  Fairon,
\newblock ``A hybrid rule/model-based finite-state framework for normalizing
  sms messages,''
\newblock in {\em Proceedings of the 48th Annual Meeting of the Association for
  Computational Linguistics}. Association for Computational Linguistics, 2010,
  pp. 770--779.

\bibitem{R12}
Md~Shad Akhtar, Utpal~Kumar Sikdar, and Asif Ekbal,
\newblock ``Iitp: Hybrid approach for text normalization in twitter,''
\newblock in {\em Proceedings of the Workshop on Noisy User-generated Text},
  2015, pp. 106--110.

\bibitem{R6}
Hao Zhang, Richard Sproat, Axel~H Ng, Felix Stahlberg, Xiaochang Peng, Kyle
  Gorman, and Brian Roark,
\newblock ``Neural models of text normalization for speech applications,''
\newblock {\em Computational Linguistics}, vol. 45, no. 2, pp. 293--337, 2019.

\bibitem{R7}
Richard Sproat and Navdeep Jaitly,
\newblock ``Rnn approaches to text normalization: A challenge,''
\newblock {\em arXiv preprint arXiv:1611.00068}, 2016.

\bibitem{R8}
Courtney Mansfield, Ming Sun, Yuzong Liu, Ankur Gandhe, and Bj{\"o}rn
  Hoffmeister,
\newblock ``Neural text normalization with subword units,''
\newblock in {\em Proceedings of the 2019 Conference of the North American
  Chapter of the Association for Computational Linguistics: Human Language
  Technologies, Volume 2 (Industry Papers)}, 2019, pp. 190--196.

\bibitem{R14}
Richard Sproat and Navdeep Jaitly,
\newblock ``An rnn model of text normalization.,''
\newblock in {\em INTERSPEECH}. Stockholm, 2017, pp. 754--758.

\bibitem{R9}
Ashish Vaswani, Noam Shazeer, Niki Parmar, Jakob Uszkoreit, Llion Jones,
  Aidan~N Gomez, {\L}ukasz Kaiser, and Illia Polosukhin,
\newblock ``Attention is all you need,''
\newblock in {\em Advances in neural information processing systems}, 2017, pp.
  5998--6008.

\bibitem{R10}
Tsung-Yi Lin, Priya Goyal, Ross Girshick, Kaiming He, and Piotr Doll{\'a}r,
\newblock ``Focal loss for dense object detection,''
\newblock in {\em Proceedings of the IEEE international conference on computer
  vision}, 2017, pp. 2980--2988.

\bibitem{R15}
Jacob Devlin, Ming-Wei Chang, Kenton Lee, and Kristina Toutanova,
\newblock ``Bert: Pre-training of deep bidirectional transformers for language
  understanding,''
\newblock 2018.

\end{thebibliography}

\end{document}